\documentclass[manuscript, nonacm, a4paper, natbib]{acmart}

\usepackage{amsmath,amsfonts,bm}

\newtheorem{theorem}{Theorem}[section]

\newtheorem{definition}[theorem]{Definition}

\def\eqref#1{equation~\ref{#1}}

\def\1{\bm{1}}

\def\va{{\bm{a}}}
\def\vb{{\bm{b}}}
\def\vc{{\bm{c}}}

\def\vf{{\bm{f}}}
\def\vg{{\bm{g}}}
\def\vh{{\bm{h}}}
\def\vi{{\bm{i}}}

\def\vo{{\bm{o}}}

\def\vr{{\bm{r}}}
\def\vs{{\bm{s}}}

\def\vu{{\bm{u}}}

\def\vw{{\bm{w}}}
\def\vx{{\bm{x}}}
\def\vy{{\bm{y}}}
\def\vz{{\bm{z}}}

\def\mA{{\bm{A}}}

\def\mD{{\bm{D}}}

\def\mF{{\bm{F}}}

\def\mH{{\bm{H}}}
\def\mI{{\bm{I}}}

\def\mK{{\bm{K}}}
\def\mL{{\bm{L}}}
\def\mM{{\bm{M}}}

\def\mO{{\bm{O}}}

\def\mQ{{\bm{Q}}}

\def\mS{{\bm{S}}}
\def\mT{{\bm{T}}}

\def\mV{{\bm{V}}}
\def\mW{{\bm{W}}}
\def\mX{{\bm{X}}}
\def\mY{{\bm{Y}}}

\DeclareMathAlphabet{\mathsfit}{\encodingdefault}{\sfdefault}{m}{sl}
\SetMathAlphabet{\mathsfit}{bold}{\encodingdefault}{\sfdefault}{bx}{n}

\def\gC{{\mathcal{C}}}
\def\gD{{\mathcal{D}}}
\def\gE{{\mathcal{E}}}
\def\gF{{\mathcal{F}}}
\def\gG{{\mathcal{G}}}

\def\gL{{\mathcal{L}}}
\def\gM{{\mathcal{M}}}

\def\gQ{{\mathcal{Q}}}

\def\gS{{\mathcal{S}}}
\def\gT{{\mathcal{T}}}
\def\gU{{\mathcal{U}}}
\def\gV{{\mathcal{V}}}
\def\gW{{\mathcal{W}}}
\def\gX{{\mathcal{X}}}
\def\gY{{\mathcal{Y}}}
\def\gZ{{\mathcal{Z}}}

\AtBeginDocument{%
  \providecommand\BibTeX{{%
    \normalfont B\kern-0.5em{\scshape i\kern-0.25em b}\kern-0.8em\TeX}}}

\begin{document}

\title{A Primer on Temporal Graph Learning}

\author{Aniq Ur Rahman}
\email{aniq.rahman@eng.ox.ac.uk}
\orcid{1234-5678-9012}
\author{Justin P. Coon}
\email{justin.coon@eng.ox.ac.uk}
\affiliation{%
  \institution{Department of Engineering Science, University of Oxford}
  \streetaddress{Parks Road}
  \city{Oxford}
  \country{UK}
  \postcode{OX1 3PJ}
}

\renewcommand{\shortauthors}{Rahman and Coon}

\begin{abstract}
  This document aims to familiarize readers with temporal graph learning (TGL) through a concept-first approach. We have systematically presented vital concepts essential for understanding the workings of a TGL framework. In addition to qualitative explanations, we have incorporated mathematical formulations where applicable, enhancing the clarity of the text. Since TGL involves temporal and spatial learning, we introduce relevant learning architectures ranging from recurrent and convolutional neural networks to transformers and graph neural networks. We also discuss classical time series forecasting methods to inspire interpretable learning solutions for TGL.
\end{abstract}

\settopmatter{printacmref=false}

\maketitle

\section{Introduction}
Graphs are expressive data structures capable of modelling intricate relationships among different entities across various domains. For example, the interaction of different users on a social media platform can be represented as graphs. Similarly, citation networks capture the scholarly interdependence of academic papers through citations, offering insights into the knowledge landscape. Biological networks, encompassing interactions among bio-molecules, genes, and proteins, leverage graphs to unravel the complexities of biological systems. 
Building upon the accomplishments of neural networks in processing Euclidean data, graph neural networks (GNNs) are specifically crafted to operate seamlessly with graph-structured data. The foundations of GNNs can be traced to graph signal processing (GSP), which was developed to impart meaning to signal processing operations on graphical data.

Temporal graphs serve as a valuable tool for modelling dynamic systems that undergo evolution over time. Conceptually, temporal graphs can be likened to time series data, where individual data points correspond to distinct graphs. The objective of temporal graph learning is to adeptly capture both the spatial dependencies within a given graph and the temporal relationships that exist with other graphs in past and future instances. GNNs effectively handle static graphs, yet they face challenges in adapting to the dynamic nature of temporal graphs. As a result, researchers have developed temporal GNNs that combine spatial learning (GNNs) and temporal learning.

The purpose of this document is to introduce temporal graph learning to the readers with a concept-first approach. We have introduced most of the concepts required to understand the workings of a temporal graph learning framework. Along with a qualitative description, we have provided mathematical formulations wherever possible to further aid the understanding of the readers.

We start off by reviewing the literature associated with temporal graph learning (TGL) in Sec.~\ref{sec:graphs}. The survey on temporal GNNs \cite{longa_graph_2023} was helpful in formalising the definitions of the learning tasks. In the next section, we introduced key concepts from graph signal processing based on the text in \cite{shuman_emerging_2013, sandryhaila_discrete_2013, ortega_graph_2018}. Then we delve into the explanation of neural networks relevant to TGL, such as recurrent neural network, convolutional neural networks, self-attention networks, variational autoencoders, and GNNs.
In Sec.~\ref{sec:classic}, we briefly  explain some classical techniques for time series forecasting to draw inspiration for developing interpretable learning solutions for TGL.
Finally, in Sec.~\ref{sec:direction}, we discuss the current limitations and potential applications of TGL highlighting the research directions of interest.


\subsubsection*{Notations} We use lower case letters to depict scalars $x$, lower case bold letters for vectors $\vx$, upper case bold letters for matrices $\mX$, and cursive upper case letters to represent sets $\gX$. The transpose of $\mX$ is $\mX^\top$, and $\mX^{\rm H}$ denoted the Hermitian of a complex matrix $\mX$. The set of natural integers $\{ 1, 2, \cdots, n \}$ is written as $[n]$. The Hadamard product is denoted by $\odot$ and the Hadamard division by $\oslash$. We label the vectors or matrices by the subscript, and use superscript to index them, for example, the $i^{\rm th}$ element of vector $\vx$ is denoted as $\vx^i$. For a matrix or tensor $\mX$, the element at the $i^{\rm th}$ row and $j^{\rm th}$ column is accessed by $\mX^{i,j}$. Moreover, a matrix $\mX$ raised to the power $k \in \mathbb{N}$ is denoted as $\mX^{(k)}$ The cardinality of a set $\gX$ is denoted by $|\gX|$.

\section{Temporal Graphs} 
\label{sec:graphs}

\subsection{Definitions}
\begin{definition}[Graph]
    Graph is a structure representing a set of inter-related  nodes where the relation between the nodes is depicted through the set of edges $\gE \subseteq \gV \times \gV$. For nodes $u, v \in \gV$ the edge connecting them is denoted by the tuple $(v,u)$ or $(u,v)$. The node features each of dimension $d \in \mathbb{N}$ are represented by a matrix $\mX \in \mathbb{R}^{|\gV|\times d}$ and the edge features are represented by a matrix $\mW \in \mathbb{R}^{|\gE| \times c}$, where $c \in \mathbb{N}$ is the dimension of each edge feature. Therefore, the entire graph can be described by the quadruple $(\gV, \gE, \mX, \mW)$.
    \label{def:graph}
\end{definition}

\begin{definition}[Adjacency matrix]
    The adjacency matrix $\mA$ of a graph $(\gV, \gE)$ with $\gV = [n]$ is defined as
    \begin{align}
        \mA^{u,v} \triangleq \mathbb{I}\{ (u,v) \in \gE \}, \quad \forall u,v \in \gV.
    \end{align}  
\end{definition}

\begin{definition}[Weighted Adjacency matrix]
    The weighted adjacency matrix of a graph $(\gV, \gE, \mW)$ is defined as
    \begin{align}
        \mA^{u,v} \triangleq \mW^{\langle u,v \rangle} \cdot  \mathbb{I}\{ (u,v) \in \gE \}, \quad \forall u,v \in \gV, \, \mW \in \mathbb{R}^{|\gE|\times c},
    \end{align} 
    where $\langle u, v \rangle : [|\gV|] \times [|\gV|] \rightarrow [|\gE|]$.
\end{definition}

\begin{definition}[$k$-hop neighbours]
    We define the $k$-hop neighbours of node $v \in \gV= [n]$ as 
    \begin{align}
    \gU_v^{[k]} \triangleq \left\{ u : \left( \mA^{(k)} \right)^{u,v} > 0, \, \forall u \in \gV \right\}.
    \end{align}
\end{definition}

In a directed graph, edge $(u, v)$ represents an arrow directed from $u$ to $v$, whereas in an undirected graph, $(v,u) \equiv (v,u)$ and represents a line connecting the nodes $u$ and $v$. The node and edge feature matrices are optional in graph definition. For clarity, 1-hop neighbours are also represented as simply $\gU_v$.

\begin{definition}[Temporal graph]
    A temporal graph is a graph that evolves over time. The set of nodes $\gV$ may change, the set of edges $\gE$ may change, or the node and edge feature matrices may change. For time stamp $t \in \gT$ the graph is denoted as $\gG_t = (t, \gV_t, \gE_t, \mX_t, \mW_t)$ and the temporal graph is represented as the chain $\gG = \{ \gG_t : \forall t \in \gT \}.$ 
    \label{def:tempgraph}
\end{definition}

\begin{definition}[Graph space]
    The domain of a temporal graph instance is referred to as the graph space, i.e., $\mathbb{G} \triangleq {\rm dom}(\gG_t)$.
\end{definition}

The elements of $\gT$ are sorted in ascending order, i.e., $\gT_i < \gT_j$ for any $j > i$. If the difference between time stamps is constant throughout $\gT$, we can replace $\gT$ with $[T]$, where $T= |\gT|$. An alternate definition for temporal graphs also exists where at each time stamp of change, only the respective changes are recorded. 

\subsection{Learning Tasks}
We can perform supervised and unsupervised learning tasks on temporal graphs. Regression, classification, link prediction, and generation are supervised, whereas clustering and low-dimensional embedding are unsupervised.

\subsubsection{Regression} In regression we relate a dependent variable to explanatory variable(s) \cite{freedman_statistical_2009}. In the context of temporal graphs, the dependent variables are the node or edge feature matrix and the explanatory variables are the temporal graphs. Therefore, the regression task is to estimate the node or edge feature matrix at a future time, given the past observations of the temporal graph.

\begin{definition}[Node regression] 
Predict the node feature matrix $\mX_t \in \mathbb{R}^{|\gV_t| \times d}$ for some $t > \tau$, given 
$\{\gG_t : \forall t \in \gT' \}$, where $\gT' \subseteq \gT \cap (\infty, \tau) $ and $\gG_t = (t, \gV_t, \gE_t, \mX_t, \mW_t )$, $\mW_t$ being optional.
\begin{align}
    f_{\rm NR}: \mathbb{G}^{T} \rightarrow  \mathbb{R}^{n \times d},  \qquad n= |\gV_t|, T = |\gT'|.
\end{align}
\end{definition}

\begin{definition}[Edge regression] 
Predict the edge feature matrix $\mW_t \in \mathbb{R}^{|\gE_t| \times c}$ for some $t > \tau$, given 
$\{\gG_t : \forall t \in \gT' \}$, where $\gT' \subseteq \gT \cap (\infty, \tau) $ and $\gG_t = (t, \gV_t, \gE_t, \mW_t, \mX_t )$, $\mX_t$ being optional.
\begin{align}
    f_{\rm ER}: \mathbb{G}^{T} \rightarrow  \mathbb{R}^{m \times d},  \qquad m = |\gE_t|, T = |\gT'|.
\end{align}
\end{definition}

\subsubsection{Classification}
In classification, the goal is to assign a class to a variable from a set of valid classes based on the explanatory variables.
In the context of temporal graphs, we have to define three graph descriptors $\vy_t, \vz_t, \vg_t$. The descriptor $\vy_t \in \gD^{|\gV_t|}$  denotes the class of each node, and $\vz_t \in \gC^{|\gE_t|}$  denotes the class of each edge. Moreover, a graph at time $t$ is assigned a class from the set $\gM$, i.e., $\vg_t \in \gM$.

\begin{definition}[Node classification] 
Predict the node class vector $\vy_t \in \gD^{|\gV_t|}$ for some $t > \tau$, given 
$\{\gG_t : \forall t \in \gT' \}$, where $\gT' \subseteq \gT \cap (\infty, \tau) $, $\gG_t = (t, \gV_t, \gE_t, \vy_t, \gQ_t )$, and the optional variables are represented by $\gQ_t = (\mX_t, \mW_t, \vz_t)$.
\begin{align}
    f_{\rm NC}: \mathbb{G}^{T} \rightarrow  \gD^n ,  \qquad n = |\gV_t|, T = |\gT'|.
\end{align}
\end{definition}

\begin{definition}[Edge classification] 
Predict the edge class vector $\vz_t \in \gC^{|\gE_t|}$ for some $t > \tau$, given 
$\{\gG_t : \forall t \in \gT' \}$, where $\gT' \subseteq \gT \cap (\infty, \tau) $, $\gG_t = (t, \gV_t, \gE_t, \vz_t, \gQ_t )$, and the optional variables are represented by $\gQ_t = (\mX_t, \mW_t, \vy_t)$.
\begin{align}
    f_{\rm EC}: \mathbb{G}^{T} \rightarrow  \gC^m ,  \qquad m= |\gE_t|, T = |\gT'|.
\end{align}
\end{definition}

\begin{definition}[Graph Classification] 
Predict the class of the graph $\vg_t \in \gM$ for some $t \geq \tau$, given 
$\{\gG_t : \forall t \in \gT' \}$, where $\gT' \subseteq \gT \cap (\infty, \tau) $, $\gG_t = (t, \gV_t, \gE_t, \gQ_t )$, and optional variables $\gQ_t = (\vg_t, \mX_t, \mW_t, \vy_t, \vz_t)$.
\begin{align}
    f_{\rm GC}: \mathbb{G}^{T} \rightarrow  \gM.
\end{align}
\end{definition}

\subsubsection{Link prediction}
In link prediction, the goal is to estimate the probability of an edge's existence at any time stamp $t$, based on past observations of the graph.
\begin{definition}[Link prediction]
    Estimate $P\left( (v,u) \in \gE_t \right) \, \forall (u, v) \in \gV_t \times \gV_t $ for some $t > \tau$, given $\{\gG_t : \forall t \in \gT' \}$, where $\gT' \subseteq \gT \cap (\infty, \tau) $, where $\gG_t = (t, \gV_t, \gE_t, \gQ_t )$, and $\gQ_t = (\mX_t, \mW_t, \vy_t, \vz_t)$.
    \begin{align}
        f_{\rm LP}: \mathbb{G}^{T} \rightarrow  [0,1]^m,  \qquad m= |\gV_t|^2, T = |\gT'|.
    \end{align}
\end{definition}

\subsubsection{Graph generation}
In the context of static graphs, graph generation refers to the task of observing a set of graphs sharing common properties, and then generating a new graph possessing similar properties. The properties and similarity metric are defined based on the task at hand. For temporal graphs,  we can generate future graphs based on past observations. Alternatively, graph generation for temporal graphs can also be defined as the task of reproducing a similar but unidentical sequence of graphs based on the observations.

\begin{definition}[Graph generation, Predictive]
    Generate a futuristic sequence of graphs $\{ \gG_t : \forall t \in \gT^* \}$ where $\gT^* \subset \mathbb{R} \cap [ \tau, \infty)$, given the past sequence $\{\gG_t : \forall t \in \gT' \}$, where $\gT' \subseteq \gT \cap (\infty, \tau)$.
    \begin{align}
        f_{\rm GG-P}: \mathbb{G}^{T} \rightarrow  \mathbb{G}^{T^*}, \qquad T = |\gT'|, T^* = |\gT^*|.
    \end{align}
\end{definition}
\begin{definition}[Graph generation, Similar]
    Given a sequence of graphs $\gG = \{ \gG_t : \forall t \in \gT' \}$ where $\gT' \subseteq \gT$, generate a sequence $\hat{\gG} = \{ \hat{\gG}_t : \forall t \in \gT' \}$ such that $\delta(\gG, \hat{\gG})$ is minimised for $\gG \neq \hat{\gG}$, where $\delta( \cdot, \cdot )$ is a distance metric.
    \begin{align}
        f_{\rm GG-S}: \mathbb{G}^{T} \rightarrow  \mathbb{G}^{T}, \qquad T = |\gT'|.
    \end{align}
\end{definition}

\subsubsection{Clustering}
Clustering can be viewed as the unsupervised counterpart of classification tasks, therefore we have to define clustering at three levels: node, edge, and graph.
\begin{definition}[Node Clustering]
    Given a set of graphs $\{ \gG_t : \forall t \in \gT' \}$ where $\gT' \subseteq \gT$, partition the nodes in each graph $\gG_t$ into $k$ clusters, where $\gG_t = (t, \gV_t, \gE_t, \gQ_t )$, and the optional variables are represented by $\gQ_t = (\mX_t, \mW_t, \vz_t)$.
    \begin{align}
        f_{\rm NCX}: \mathbb{G}^{T} \rightarrow [k]^{n \times T},  \qquad n, T = |\gT'|.
    \end{align}
\end{definition}
\begin{definition}[Edge Clustering]
    Given a set of graphs $\{ \gG_t : \forall t \in \gT' \}$ where $\gT' \subseteq \gT$, partition the edges in each graph $\gG_t$ into $k$ clusters, where $\gG_t = (t, \gV_t, \gE_t, \gQ_t )$, and the optional variables are represented by $\gQ_t = (\mX_t, \mW_t, \vy_t)$.
    \begin{align}
        f_{\rm ECX}: \mathbb{G}^{T} \rightarrow [k]^{m \times T},  \qquad m, T = |\gT'|.
    \end{align}
\end{definition}
\begin{definition}[Graph Clustering]
    Partition a set of graphs $\{ \gG_t : \forall t \in \gT' \}$ where $\gT' \subseteq \gT$, into $k$ clusters, where $\gG_t = (t, \gV_t, \gE_t, \gQ_t )$, and the optional variables are represented by $\gQ_t = (\mX_t, \mW_t, \vy_t, \vz_t)$.
    \begin{align}
        f_{\rm GCX}: \mathbb{G}^{T} \rightarrow [k]^{T}, \qquad T = |\gT'|.
    \end{align}
\end{definition}

\subsubsection{Low-dimensional embedding}
Low-dimensional embedding (LDE) refers to the compression of a graph by representing it in a lower dimension. The compression may be lossy or lossless based on the design of the encoder.

\begin{definition}[Low-dimensional graph embedding]
    Given a set of graphs $\{ \gG_t : \forall t \in \gT' \}$ where $\gT' \subseteq \gT$, the goal is to encode each graph $\gG_t$ in $\mathbb{R}^{L}$ space. 
    \begin{align}
        f_{\rm LDE-G} : \mathbb{G} \rightarrow \mathbb{R}^L,  \qquad L \leq  |\gV_t|(d+1) + |\gE_t|(c+2).
    \end{align}
\end{definition}

\begin{definition}[Low-dimensional graph sequence embedding]
    Given a set of graphs $\gG = \{ \gG_t : \forall t \in \gT' \}$ where $\gT' \subseteq \gT$, encode the entire graph sequence $\gG$ in $\mathbb{R}^{L}$ space. 
    \begin{align}
        f_{\rm LDE-GS} : \mathbb{G}^{T} \rightarrow \mathbb{R}^L,  \qquad L \leq \sum_{t \in \gT'} \Big(  |\gV_t|(d+1) + |\gE_t|(c+2) \Big), T = |\gT'|.
    \end{align}
\end{definition}

Graph compression is lossless only if the embedding function is invertible.

\subsection{Output type}
The learning tasks described above can either be deterministic or probabilistic \cite{lim_time_2021}.
\subsubsection{Probabilistic output}
Since predictions involve some level of uncertainty, the realistic approach is to return a value along with its likelihood. In the context of learning, we need to estimate the distribution of the future output. A general approach is to approximate the data to a standard probability distribution, such as normal, uniform, or beta, or a Gaussian mixture model whose parameters are learnt based on the observations.

\subsubsection{Deterministic output}
In the case that the output is deterministic, the goal is to learn the expected future value of the required variable, i.e., we learn the mean estimator of the future variable. While such approaches result in the least mean square error (LMSE) in the training data, they fail to give us an idea of the confidence in the output.

\section{Graph Signal Processing}
Typical signal processing tasks such as translation, modulation, filtering, sampling, and convolution could not be readily applied to non-Euclidean data such as graphs. Therefore, graph signal processing (GSP) was developed as a mathematical tool that gave intuitive meaning to signal processing tasks on graphical data \cite{shuman_emerging_2013, sandryhaila_discrete_2013, ortega_graph_2018}.
For the remainder of this section. We consider graphs as defined in \textit{def.}~\ref{def:graph} with $\gV=[n]$, $\mX \in \mathbb{C}^n$, and weighted adjacency matrix $\mA \in \mathbb{C}^{n \times n}$.

\begin{definition}[Degree matrix]
    The degree of a node in a graph is defined as the count of nodes connected to that node. Consequently, the degree matrix is defined as $\mD = {\rm diag}( \bm{1}^\top \mA  )$, where $\mD_{v,v}$ represents the sum of the weights of the edges connected to the node $v \in \gV$.
\end{definition}

\begin{definition}[Graph Laplacian matrix]
    The graph Laplacian matrix is defined as $\mL \triangleq \mD - \mA$.
\end{definition}

\begin{definition}[Normalized graph Laplacian matrix]
    The normalized graph Laplacian matrix is defined as $\hat{\mL} \triangleq \mD^{-1/2} \mL \mD^{-1/2}$.
\end{definition}

In digital signal processing (DSP), the concept of shift or delay is used to represent a signal of $n$ samples. Let $z^{-1}$ be a shift, then the signal $\vs = \{ s_0, s_1, \cdots s_{n-1}  \}$ is represented as the sum $S(z) = \sum_{k=0}^{n-1} s_k z^{-k}$, also known as $z$-transform. There exists an inverse operation, called the inverse $z$-transform through which, we can recover the signal $\vs$, given $S(z)$. Extending the concept of shifts and transforms to GSP, we consider a square matrix $\mS \in \mathbb{C}^{n \times n}$ as the shift; $\mS = \mV \bm{\Lambda} \mV^{-1}$ where $\mV$ is the matrix of $n$ eigenvectors of $\mS$, and $\bm{\Lambda}$ is a diagonal matrix consisting of the distinct eigenvalues of $\mS$. The shift for a graph can be chosen to be $\mA$, $\mL$ or $\hat{\mL}$, or something else by design \cite{ortega_graph_2018}.

\begin{definition}[Characteristic Polynomial]
    The characteristic polynomial of $\mA$ is a polynomial the roots of which are the eigenvalues of $\mA$, i.e., $ p_{\mA}(\vz) = \det(\mA - \vz \mI)$.
\end{definition}

\begin{definition}[Shift-invariant filter]
    A shift-invariant filter $\mH$ is a polynomial $h(\cdot)$ in the shift $\mS$, i.e. $\mH = h(\mS)$ such that ${\rm degree}(h(z)) \leq {\rm degree}(p_{\mS}(z)) \leq n -1 $, where $p_{\mS}(z)$ is the characteristic polynomial of $\mS$.
\end{definition}

\begin{definition}[Graph Fourier transform]
    The graph Fourier transform operator $\mF$ is defined as the inverse of the matrix of the eigenvectors of the shift $\mS$, i.e., $\mF = \mV^{-1}$. The graph Fourier transform on a graph signal $\vx \in \mathbb{C}^n$ is defined as $\gF(\vx) \triangleq \mF \vx$, and the inverse graph Fourier transform is defined as $\gF^{-1}(\vy) \triangleq \mF^{-1} \vy$, where $\vy$ is a signal in the graph frequency domain.
\end{definition}

For $\mS = \hat{\mL}$, $\mV^{-1} = \mV^{\rm H}$, therefore, $\gF(\vx) = \mV^{\rm H} \vx$ and $\gF^{-1}(\vy) = \mV \vy$.

\begin{definition}[Graph convolution]
    Given the shift $\hat{\mL}$, a filter $\vg \in \mathbb{C}^n$ and graph signal $\vx \in \mathbb{C}^n$, we define the graph convolution operation as $\vx \star_{\gG} \vg \triangleq \gF^{-1} \Big( \gF(\vx) \odot \gF(\vg) \Big)$, where $\star_{\gG}$ denotes the graph convolution operator. If by design $\vg_{\theta} = {\rm diag}(\mV^{\rm H} \vg)$, the expression of the convolution operation simplifies to $\vx \star_{\gG} \vg_{\theta} = \mV \vg_{\theta} \mV^{\rm H} \vx$.
    \label{def:graphconv}
\end{definition}


\section{Relevant Neural Networks} 
In this section, we will describe the neural networks relevant to graph learning \cite{jin_spatio-temporal_2023, wu_comprehensive_2021} and temporal learning \cite{lim_time_2021}, and explain how the fusion of the two is used for TGL \cite{longa_graph_2023}.

\subsection{Recurrent Neural Networks}
A recurrent neural network (RNN) is a feed-forward neural network capable of taking variable-length sequences as input \cite{chung_empirical_2014}. It does so, by maintaining a hidden state at each time step which is derived from a function of the previous hidden state and current input. The hidden state encodes the historical information of the sequence \cite{lim_time_2021}.
\begin{definition}[RNN]
    Given an input sequence $(\vx_1, \vx_2, \cdots \vx_T)$, the hidden state at time $t$, $\vh_t$ is given by the recursive expression $\vh_t = f(\vh_{t-1}, \vx_t)$. Then, the conditional distribution is approximated by the function $g(\vh_t) = p(\vx_t | \vx_{t-1}, \cdots \vx_1)$. The learnable functions $f(\cdot, \cdot)$ and $g(\cdot)$ constitute the RNN.
    \label{def:RNN}
\end{definition}

\subsubsection{Long-Short Term Memory}
Long-Short Term Memory (LSTM) was proposed by \cite{hochreiter_long_1997} to overcome the difficulty of conventional RNNs in learning long-range data dependencies \cite{lim_time_2021}.
The hidden state $\vh_t$ is updated through a series of gates (forget: $\vf_t$, input: $\vi_t$, and output: $\vo_t$), designed to modulate the historical content encoded in the cell state $\vc_t$.
The architecture is also capable of tackling gradient vanishing and explosion. Each of the gates $\vf_t, \vi_t, \vo_t$ is of the form $[\cdot]_t = \sigma( \mW_{[\cdot]} [ \vh_{t-1} \quad \vx_t ] + \vb_{[\cdot]} )$ where $\sigma(\cdot)$ is the sigmoid activation function. The candidate cell state uses tanh activation, $\hat{\vc}_t = {\rm tanh}(\mW_c [ \vh_{t-1} \quad \vx_t ] + \vb_c )$. The hidden state is updated as
\begin{align}
    \vh_t = \vo_t \odot {\rm tanh}( \vc_t ); \quad \vc_t = \vf_t \odot \vc_{t-1} + \vi_t \odot \hat{\vc}_{t}.
\end{align}

\subsubsection{Gated Recurrent Unit}
An alternate gated RNN called Gated Recurrent Unit (GRU) was proposed by \cite{cho_properties_2014} where the recurrent unit can capture the data dependencies across different time scales \cite{chung_empirical_2014}. It consists of two gates, update $\vu_t$ and reset $\vr_t$, of the form $[\cdot]_t = \sigma( \mW_{[\cdot]} [ \vh_{t-1} \quad \vx_t ] + \vb_{[\cdot]} )$. The candidate hidden state uses tanh activation, $\hat{\vh}_t = {\rm tanh}(\mW_h [ \vr_t \odot \vh_{t-1} \quad \vx_t ] + \vb_h )$. Finally, the hidden state is updated as the linear interpolation between $\vh_{t-1}$ and $\hat{\vh}_t$ based on the update gate $\vu_t$, i.e., $\vh_t = \vu_t \odot \vh_{t-1} + (\bm{1} - \vu_t) \odot \hat{\vh}_t$.

\subsection{Convolutional Neural Networks}
The Convolutional Neural Network (CNN) stands out as a specialised neural network architecture adept at capturing local spatial dependencies within multi-dimensional data \cite{lim_time_2021}. This attribute makes CNNs prevalent in computer vision as these networks can learn the relationship between nearby pixels. 

The CNN architecture can be described as a sequence of the following layers \cite{oshea_introduction_2015}: (1) input layer, (2) convolutional layer, (3) rectified linear unit (ReLU), (4) pooling later, (5) fully-connected neural network layers with appropriate activation, (6) output layer.

\begin{definition}[$n$-Dimensional Convolution]
    Consider an $n$-dimensional matrix $\mX \in \mathbb{R}^{d_1 \times d_2 \cdots \times d_n}$, and an $n$-cubic kernel $\bm{\Theta} \in \mathbb{R}^{d \times \cdots \times d}$. The convolution $\mY = \bm{\Theta} \star \mX$ results in $\mY \in \mathbb{R}^{\bigtimes_{i=1}^n (d_i -d +1) }$ defined as:
    \begin{align}
        \mY_{i_1, \cdots, i_n} = \bm{1}^\top \left( \bm{\Theta} \odot \mX_{i_1:i_1+d, \cdots, i_n:i_n+d} \right) \bm{1}.
    \end{align}
\end{definition}

A kernel can be likened to a discernible pattern, and the process of convolution involves scanning the data to identify where this pattern is present. In CNNs, an array of trainable kernels is employed to detect various patterns within the data, based on which the downstream task is performed.

\begin{definition}[Pooling]
    In the pooling process, the data matrix $\mX$ is partitioned into regular tiles consisting of multiple units, and within each tile, the data is aggregated using a specific rule into a single unit. The resulting representations are then arranged in the same order as the tiles, producing a downsampled version of $\mX$.
\end{definition}

The pooling layer reduces the dimensionality of its input thereby reducing the number of learnable parameters required in the successive layers. The pooling layer can be viewed as a downsampler which operates based on a rule. Some common pooling layers are (1) max-pooling, (2) average-pooling, and (3) $\ell$-normalisation.

An $n$-cubic kernel of width $d$ is capable of summarising the information present in the volume of width $d$, thereby performing $d^n$ multiplication operations for each index $(i_1, \cdots, i_n)$. To make the convolution operation faster, we can reduce the width of the kernel but that reduces the reception field of the kernel. In dilated convolution, by setting certain kernel parameters to zero at regular intervals along each dimension, we create a dilation pattern resembling an $n$-dimensional lattice. This structural spacing allows the kernel to maintain a larger receptive field, capturing information from a broader region, while lowering the computational cost by reducing the number of active parameters in the kernel.

\begin{definition}[Dilated Convolution]
    Dilated convolution with dilation factor $l$ is denoted as $\mY = \bm{\Theta} \star_l \mX$, where 
    $\bm{\Theta}$ is of the form $\bm{\Theta} = \mM \odot \bm{\Theta}'$; $\bm{\Theta}'$ being an $n$-cube of width $d$ in the real space, and $\mM$ is the $l$-dilation mask with $\mM_{i_1, \cdots, i_n} = \mathbb{I}\left\{  \big[ i_k \mod l : \forall k \in [n] \big] = \bm{0}_n \right\}, \, \forall (i_1, \cdots, i_n) \in [d]^n$.
\end{definition}
In simple words, the $l$-dilation mask is one at indices where each dimension's index is an integer multiple of $l$, and zero everywhere else. This creates a lattice of ones and zeros, and out of the $d^n$ parameters in the kernel, $d^n - \left\lfloor \frac{d}{l} \right\rfloor^n$ parameters become zero.

\subsection{Temporal Convolutional Networks}
Temporal convolutional network (TCN) is a 1D-CNN where the convolution is performed along the time axis \cite{jin_spatio-temporal_2023}. The activation $\vz$ for input $\vx$ can be written as $\vz=  a( \bm{\Theta}_f \star \vx )$, where $a(\cdot)$ is the activation function and $\bm{\Theta}_f$ is the learnable convolution kernel. If the activation function is linear, then a single-layer TCN is reduced to an auto-regressive model \cite{lim_time_2021}. Stacking multiple convolutional layers facilitates context-aggregation which is necessary for temporal learning.

\subsubsection{Gated TCN}
Integrating gated RNN with TCN results in Gated TCN \cite{dauphin_language_nodate}, which improves temporal learning. In a Gated-TCN we have two different 1D-CNNs fused together. The activation $\vz$ for the input $\vx$ is given as 
\begin{align}
    \vz=  {\rm tanh}( \bm{\Theta}_f \star \vx ) \odot \sigma( \bm{\Theta}_g \star \vx ),
\end{align}
where ${\rm tanh}( \bm{\Theta}_f \star \vx )$ represents the filter and $\sigma( \bm{\Theta}_g \star \vx )$ represents the gate, $\{ \bm{\Theta}_f, \bm{\Theta}_g \}$ being learnable kernels.

\subsubsection{Causal TCN}
A regular TCN can disrupt data causality by linking past data with future data. To prevent this, Causal TCN was proposed in WaveNet \cite{oord_wavenet_2016} in which connections between past and future time-steps are explicitly dropped. Moreover, dilated convolution is used which allows the model to capture temporal patterns over a long range. The activation $\vz$ for the input $\vx$ and global conditioning variable $\vh$ is
\begin{align}
    \vz= {\rm tanh}\big( \bm{\Theta}_f \star_l \vx + \bm{\Phi}_f \star_l \vh \big) \odot \sigma\big( \bm{\Theta}_g \star_l \vx + \bm{\Phi}_g \star_l \vh \big),
\end{align}
where $\{ \bm{\Theta}_f, \bm{\Theta}_g, \bm{\Phi}_f, \bm{\Phi}_g \}$ are learnable kernels and $\star_l$ denoted dilated convolution with dilation factor $l$.

\subsection{Temporal Self-Attention Networks}
The sequential nature of processing in RNNs prevents parallelization. To alleviate this problem, an encoder-decoder architecture called Transformer \cite{vaswani_attention_2017} was proposed which uses a so-called attention mechanism to capture the global dependencies between the input and output. An attention function is a mapping from a set of key $\mK$, value $\mV$, and query $\mQ$ to an output $\mO$. In Transformers, the scaled dot-product attention (SDPA) is calculated as
\begin{align}
    f_{\rm SDPA}(\mQ, \mK, \mV) = {\rm softmax} \left( \mQ \mK^\top /\sqrt{d_k}  \right) \mV,
\end{align}
where  $\mQ \mK^\top$ computes the similarity of the query with the key, which is then scaled by a factor of $1/\sqrt{d_k}$. Transformers also use multi-head attention (MHA) which allows processing information from different sources jointly. The MHA function is defined as 
\begin{align}
    f_{\rm MHA}(\mQ, \mK, \mV, \bm{\Theta}^{O}) = \begin{bmatrix} \vh_1 & \cdots  &\vh_n   \end{bmatrix} \bm{\Theta}^{O}, \quad \vh_j = f_{\rm SDPA}\left( \mQ \bm{\Theta}_j^{Q} , \mK \bm{\Theta}_j^{K}, \mV \bm{\Theta}_j^{V} \right),
\end{align}
where $\bm{\Theta}^{[\cdot]}$ denotes the parameter matrices.
Since the transformer lacks a recurrence or convolution operator  \cite{jin_spatio-temporal_2023}, the positional information of the data in the sequence has to be explicitly embedded through Positional Encoding. 

\subsection{Variational Autoencoders}
The Variational Autoencoder (VAE) was proposed in \cite{kingma_auto-encoding_2013}, and serves as a powerful tool for generative modeling. The VAEs cab be implemented using neural networks, and the parameters can be learnt through stochastic gradient descent (SGD) methods.

\begin{definition}[Generative Model]
    Given the observation set $\gX \subset \mathbb{D}$, a generative model generates a set $\hat{\gX} \subset \mathbb{D}$, where the domain $\mathbb{D}$ is unknown.
\end{definition}

Consider a set of data points $\gX$, and a set of latent variables $\gZ$. It is assumed that the latent variables can be mapped \cite{doersch_tutorial_2021} to the data points through a generative model $f: \gZ \times \bm{\Theta} \rightarrow \gX$, where $\bm{\Theta}$ is the parameter space. In other words, there exists a parametric distribution $p_{\theta}( \vx \mid \vz)$ which can describe the likelihood, where $\vx \in \gX, \vz \in \gZ, \theta \in \bm{\Theta}$. The true posterior density can be written as:
\begin{align}
    p_{\theta}(\vz \mid \vx) = \frac{p_{\theta}(\vx \mid \vz) p_{\theta}(\vz)}{p_{\theta}(\vx)}.
\end{align}
In reality, the true posterior density is often intractable and therefore a recognition model $q_{\phi}(\vz \mid \vx)$ is used as an approximation. In a VAE, the recognition model parameters $\phi$ and the generative model parameters $\theta$ are learnt jointly by maximizing the lower bound on marginal likelihood on the observed data points $\vx \in \gX$.
\begin{align}
     \gL(\phi, \theta, \vx) \triangleq \mathbb{E}_{\vz \sim q_{\phi}(\vz \mid \vx)}\left[ \log p_{\theta}(\vx \mid \vz) \right] + D_{\rm KL}\left[ q_{\phi}(\vz \mid \vx)\, ||\, p_{\theta}(\vz)  \right],
\end{align}
where the first term is the reconstruction loss, and the second term is the KL divergence.
\begin{align}
    (\phi^*, \theta^*) = \arg \max_{\phi, \theta} \frac{1}{|\gX|}\sum_{\vx \in \gX} \gL(\phi, \theta, \vx).
\end{align}

Instead of sampling $\vz$ from $q_{\phi}(\vz \mid \vx)$ directly, the following approximation is used. The latent vector is generated from a deterministic mapping $\vz = g_{\phi}(\vx, \bm{\epsilon})$ where $\bm{\epsilon} \sim p(\bm{\epsilon})$ is an non-parameterized random variable. This is referred to as the \textit{reparameterization trick}.

\section{Graph Neural Networks}
The Graph Neural Network (GNN) was created to facilitate deep learning on graphical data, addressing the limitation of traditional deep learning architectures that are not suitable for non-Euclidean data, such as graphs. Inspired by CNN, in which a pixel is processed in unison with the adjacent pixels through convolutional kernels, the GNNs process nodes together with their neighbouring nodes to capture the spatial dependencies in the graph data. The GNNs can be broadly classified into four types \cite{wu_comprehensive_2021}: Recurrent GNN, Convolutional GNN, Graph Autoencoder, and Temporal GNN.

For the rest of this section, we employ the graph definition $\gG = (\gV, \gE, \mX, \mW)$ (see \textit{def.}~\ref{def:graph}) where the node feature vector for node $v \in \gV$ is $\mX^v \in \mathbb{R}^d$ and the weight of the edge $(v,u) \in \gE$ is $\mW^{(v,u)}$. 

\subsection{Recurrent GNN}
Recall RNN (\textit{def.}~\ref{def:RNN}), where the hidden state is updated as $\vh_t = f( \vh_{t-1}, \vx_t )$. In Recurrent Graph Neural Network (RecGNN), we follow a similar approach, but with a crucial addition. Alongside the node feature and hidden state of a node $v \in \gV$, we incorporate the node feature, hidden states, and edge weights of its neighbours $\gU_v$. Therefore, to update the hidden state of node $v$, we can write
\begin{align}
    \vh_t^{v} = f\Big( \vh_{t-1}^{v}, \mX^{v}, \{ \vh_{t-1}^{u}, \mX^{u}, \mW^{(v,u)} : \forall u \in \gU_v  \} \Big).
\end{align}

The early versions of RecGNNs were based on information diffusion, where the nodes exchange information with their neighbours recurrently until an equilibrium is reached \cite{wu_comprehensive_2021}. The update function in \cite{scarselli_graph_2009} is defined as: 
\begin{align}
    \vh_t^v = \sum_{u \in \gU_v} f \left( \mX^v, \mX^u, \mW^{(v,u)}, \vh_{t-1}^u \right).
\end{align}
The Graph Echo State Network (Graph ESN) extends the architecture of echo state network (ESN), making it compatible with graphs \cite{gallicchio_graph_2010, micheli_discrete-time_2022}. The node representation vector (hidden state) is updated through the following rule:
\begin{align}
    \vh_t^v =f_{\rm ESN} \left( \mW^{(v,v)}\mX^v + \sum_{u \in \gU_v} \mW^{(v,u)}\vh_{t-1}^u \right).
\end{align}

In a Gated GNN \cite{li_gated_2017}, the node representation vector, is updated through a GRU  which is fed the representation vectors of the node and its neighbours scaled by the edge weights. The representation vector of node $v\in \gV$ is given by
\begin{align}
    \vh_t^v = f_{\rm GRU}\left( \vh_{t-1}^v, \sum_{u \in \gU_v} \mW^{(v,u)} \vh_{t-1}^{u}  \right),
\end{align}
where $\vh_0^v = [ (\mX^v)^\top \, \bm{0} ]^\top$. After $T$ update iterations, the graph level representation vector is defined as:
\begin{align}
    \vh_{\gG} \triangleq {\rm tanh}\left( \sum_{v \in \gV} \sigma \big(g_{\theta}(\vh_T^v, \mX^v) \big) \odot {\rm tanh}\big(g_{\psi}(\vh_T^v, \mX^v)\big)   \right),
\end{align}
where $g_\theta(\cdot, \cdot)$ and $g_\psi(\cdot, \cdot)$ are neural networks that encode the inputs to a real vector.

\subsection{Convolutional GNN}
A Convolutional GNN (ConvGNN) adapts the convolution operation in Euclidean space (grid of pixels), to non-Euclidean space (graphs) \cite{wu_comprehensive_2021}. Unlike RecGNNs, where the node's representation is updated over multiple iterations until convergence, ConvGNNs operate in layers, where the node representation is updated as it propagates through the layers. RecGNNs can be seen as a special case of ConvGNNs where all the convolution layers are identical.
Like any GNN, the node representation vector in ConvGNNs is a function of its own features and the features of the neighbouring nodes.


\begin{definition}[Chebyshev polynomial]
A Chebyshev polynomial is of the form $\sum_{k=0}^{K} \bm{\alpha}_k \mT_k(\mX)$, where $(\bm{\alpha}_k, \mT_k(\mX) )$ is the $k^{\rm th}$ coefficient, exponent pair of the polynomial.
    The $k^{\rm th}$ exponent of $\mX \in \mathbb{C}^{n \times n}$ is derived from the recurrence relation: $\mT_k(\mX) = 2 \mX \mT_{k-1}(\mX) - \mT_{k-2}(\mX)$, with $\mT_0(\mX) = \mI$ and $\mT_1(\mX) = \mX$.
\end{definition}

In ChebNet \cite{defferrard_convolutional_2016}, the filter  is defined as the Chebyshev polynomial of $\Tilde{\bm{\Lambda}}$, i.e., $\vg_{\theta} = \sum_{k=0}^{K} \bm{\theta}_k \mT_k(\Tilde{\bm{\Lambda}})$, where $ \Tilde{\bm{\Lambda}} = \frac{2}{\lambda_{\max}} \bm{\Lambda} - \mI$, and $\bm{\Lambda}$ is the matrix of eigenvalues of the normalized graph Laplacian matrix.
Following \textit{def.}~\ref{def:graphconv}, the graph convolution in ChebNet is defined as
\begin{align}
    \vx \star_{\gG} \vg_{\theta} = \mV \left( \sum_{k=0}^{K} \bm{\theta}_k \mT_k(\Tilde{\bm{\Lambda}})  \right)\mV^{\rm H} \vx.
\end{align}
Through first-order approximation of ChebNet, we get $\vx \star_{\gG} \vg_{\theta} = \bm{\theta}_0 \vx - \bm{\theta}_1 \mD^{-\frac{1}{2}} \mA \mD^{-\frac{1}{2}} \vx$. Subsequently setting $\bm{\theta}_0 = -\bm{\theta}_1 = \bm{\theta}$, we get the graph convolution operation as defined in graph convolutional network (GCN) \cite{kipf_semi-supervised_2017}:
\begin{align}
    \vx \star_{\gG} \vg_{\theta} = \bm{\theta}  \Big( \mI +  \mD^{-\frac{1}{2}} \mA \mD^{-\frac{1}{2}} \Big) \vx = \bm{\theta} \Tilde{\mA} \vx.
\end{align}
Furthermore, to enable multiple input and output channels, the convolution operation can be rewritten as the layer 
\begin{align}
    \mH = \mX \star_{\gG} \vg_{\Theta} = a\Big( \Tilde{\mA} \mX \bm{\Theta} \Big),
\end{align}
where $a(\cdot)$ is the activation function. We can notice that, for a given node $v \in \gV$, 
\begin{align}
    \mH^v = a\left( \bm{\Theta}^\top \left( \Tilde{\mA}_{v,v} \mX^v + \sum_{u \in \gU_v} \Tilde{\mA}_{v,u} \mX^u  \right)  \right).
\end{align}
Such representations are called spatial, whereas expressions grounded in the theory of GSP are classified as spectral. 

Diffusion CNN \cite{atwood_diffusion-convolutional_2016} defines diffusion graph convolution as 
\begin{align}
    \mH_k = a\left( \bm{\Theta}_k \odot (\mD^{-1} \mA)^k \mX  \right),
\end{align}
with the final representation being $\begin{bmatrix}
    \mH_1 & \cdots & \mH_K
\end{bmatrix}$, where $a(\cdot)$ is the activation function. 
Alternatively, the final representation can be derived through summation~\cite{li_diffusion_2018} instead of concatenation of the representation at different diffusion steps, i.e., 
$\mH = \sum_{k=0}^{K} a \left( (\mD^{-1} \mA)^k \mX \bm{\Theta}_k \right)$.

In generalizing the diffusion process as information exchange between nodes in a finite number of iterations, we get message-passing neural network (MPNN) \cite{gilmer_neural_nodate}. The hidden representation of node $v$ at the $k^{\rm th}$ iteration is given by 
\begin{align}
    \vh_k^v = f\left( \vh_{k-1}^v , \sum_{u \in \gU_v} g \Big( \vh_{k-1}^v, \vh_{k-1}^u, \mA_{v,u} \Big)  \right)   
\end{align}
with $\vh_0^v = \mX^v$, where $f(\cdot), g(\cdot)$ are learnable functions.
The global graph representation is obtained by passing $\{ \vh_K^v : \forall v \in \gV \}$ to the learnable readout  function $f_R(\cdot)$, i.e., $\vh_{\gG} = f_R(\{ \vh_K^v : \forall v \in \gV \})$. It was found that MPNN architecture resulted in similar graph representations for graphs having different structures. To overcome this, the graph isomorphism network (GIN) \cite{xu_how_2019} was developed, which performs graph convolution as
\begin{align}
    \vh_k^v = f_{\rm MLP}\left( (1 + \epsilon_k) \vh_{k-1}^v + \sum_{u \in \gU_v} \vh_{k-1}^u \right),
\end{align}
where $\{ \epsilon_k: \forall k \in [K] \}$ are learnable parameters, and $f_{\rm MLP}(\cdot)$ denotes multilayer perceptron (MLP).
Since $|\gU_v| < n$ for any node $v$, the computation complexity can explode for a graph which is densely connected. To prevent this, we can sample a subset of neighbours $\gS_v \subset \gU_v$ where $|\gS_v| \leq q < n$ is constrained \cite{hamilton_inductive_nodate}. 
In the graph attention network (GAT) \cite{velickovic_graph_2018}, the weight of the edge between a node and its neighbours is made learnable, 
\begin{align}
    \vh_k^v = \sigma \left( \sum_{u \in \gU_v \cup \{ v \}} \bm{\alpha}_k^{(v,u)} \bm{\Theta}_k \vh_{k-1}^u \right),
\end{align}
where $\bm{\alpha}_k^{v} \triangleq {\rm softmax}( [ e_k^{(v,u)} : \forall u \in \gU_v \cup \{ v \} ] )$, and $e_k^{(v, u)} \triangleq f_{\rm LeakyReLU}\left( \va^\top [\bm{\Theta}_k \vh_k^v \quad \bm{\Theta}_k \vh_k^u ] \right)$. In GAT, the learnable parameters are $\va$ and $\bm{\Theta}_k$.

\subsection{Graph Autoencoder}
As implied by its name, the Graph Autoencoder (GAE) is an autoencoder  designed for graphs. GAE operates on the principle of VAEs with the primary objective of acquiring low-dimensional graph representations \cite{kipf_variational_2016, komodakis_graphvae_2019}. This is achieved by encoding the graph's structural characteristics into a latent space and subsequently decoding them to reconstruct the original graph. GAEs find utility in an array of tasks, including graph generation, and graph clustering.

\begin{definition}[Graph autoencoder]
    Let $\vz \in \mathbb{R}^L$ be the representation of a graph $\gG \in \mathbb{G}$ in the latent space. Then, we define a pair of encoder and decoder functions $f_{\rm E}: \mathbb{G} \rightarrow \mathbb{R}^L$,  $f_{\rm D}: \mathbb{R}^L \rightarrow  \mathbb{G}$; $\vz = f_{\rm E}(\gG)$, and $\hat{\gG} = f_{\rm D}(\vz)$. The goal is to design the encoder-decoder pair such that the condition on a metric involving  $\gG, \hat{\gG}, \vz$ is satisfied.
\end{definition}


In GraphVAE \cite{komodakis_graphvae_2019}, the encoder is defined as the variational posterior $q_{\phi}(\vz \mid \gG)$, and the decoder is the generative distribution $p_{\theta}(\gG \mid \vz)$. The distributions are learnt by minimizing the upper bound $\gL$ on the negative log-likelihood $- \log p_{\theta}(\gG)$.
\begin{align}
    \gL(\phi, \theta, \gG) = \mathbb{E}_{\vz \sim q_{\phi}(\vz \mid \gG)}\left[ - \log p_{\theta}(\gG \mid \vz)  \right] + D_{\rm KL}\left[ q_{\phi}(\vz \mid \gG) || p(\vz)  \right].
\end{align}  
The term $\mathbb{E}_{\sim q_{\phi}}\left[ - \log p_{\theta} \right]$ is the reconstruction loss which measures the difference between the input and the output graphs, where the difference metric is defined for the graph generation task at hand.

In \cite{guo_interpretable_2020}, authors use disentangled representation learning (DRL) for graph generation. They consider the graph $\gG = (\mX, \mW)$ to be generated from the set of latent variables $\vz_x, \vz_w, \vz_g$ such that (1) $\mX$ can be inferred from $\vz_x, \vz_g$, and (2) $\mW$ can be inferred from $\vz_w, \vz_g$. They developed an unsupervised generative model NED-VAE capable of learning the the joint distribution of the graph $(\gX, \gW)$ and the set of latent variables $\vz_x, \vz_w, \vz_g$. Utilizing a VAE architecture, the encoder is formulated as $q_{\phi}(\vz_x, \vz_w, \vz_g \mid \mX, \mW)$, and the decoder is formulated as $p_{\theta}(\mX, \mW \mid \vz_x, \vz_w, \vz_g )$. 
To learn the distributions, the following optimization problem is solved:
\begin{align}
    &\max_{\phi, \theta} \, \mathbb{E}_{(\mX, \mW) \sim \gD}\left[ \mathbb{E}_{(\vz_x, \vz_w, \vz_g) \sim q_{\phi}(\vz \mid \mX, \mW )} \, \log p_{\theta}(\mX, \mW \mid \vz_x, \vz_w, \vz_g ) \right]\\
    &\,{\rm s.t.} \quad D_{\rm KL}\left[ q_{\phi}(\vz_x, \vz_w, \vz_g \mid \mX, \mW) || p(\vz_x, \vz_w, \vz_g)   \right] < \epsilon.
\end{align}
Due to DRL, the distributions $q_{\phi}$ and $p_{\theta}$ can be disentangled into conditionally independent components:
\begin{align}
    q_{\phi}(\vz_x, \vz_w, \vz_g \mid \mX, \mW) &= q_{\phi}(\vz_x \mid \mX) \, q_{\phi}(\vz_w \mid \mW) \, q_{\phi}(\vz_g \mid \mX, \mW),\\
    p_{\theta}(\mX, \mW \mid \vz_x, \vz_w, \vz_g ) &= p_{\theta}(\mX \mid \vz_x, \vz_g) \, p_{\theta}(\mW \mid \vz_w, \vz_g).
\end{align}

Generative models designed for Euclidean data are not capable of identifying different permuations of the same graph as isomorphic, leading to poor model generalisation from the training data. In \cite{niu_permutation_2020}, the authors adopt score-based generative model \cite{song_generative_2019} and propose a permutation equivariant score function (\textit{def.}~\ref{def:score}) capable of learning permutation invariant representations of the graph.

\begin{definition}[Score function]
    For a probability distribution $p(\vx)$, the score function is defined as $\nabla_{\vx} \log p(\vx)$.
    \label{def:score}
\end{definition}

\subsection{Temporal Graph Neural Network} 
The basic idea of a temporal graph neural network (TGNN) \cite{longa_graph_2023, jin_spatio-temporal_2023} is to do with graphs what a temporal neural network does to Euclidean data, such as matrices. Let the sequence of graphs be $\{ \gG_1, \cdots, \gG_t, \cdots \gG_T \}$. At time $t$, we want to generate a representation vector $\vh_t$ which can \textit{encode} information from the sub-sequence $\{ \gG_1, \cdots, \gG_t \}$. We can write the representation vector at time $t$ as a function of all the graphs up to time $t$ and the previous representation vectors, 
\begin{align}
    \vh_t = f( \gG_t, \cdots, \gG_1 , \vh_{t-1}, \cdots, \vh_1  ).
\end{align}
Assuming that the information from previous graphs will propagate through the previous representation vectors, we can write $\vh_t = f(\gG_t, \vh_{t-1} ,\cdots \vh_1)$. Furthermore, we can decompose the function into $\vh_t = f_{\rm T}\big( f_{\rm S}(\gG_t), \vh_{t-1}, \cdots \vh_1 \big)$, where $f_{\rm T}$ is a temporal learning function and $f_{\rm S}$ is a spatial (graph) learning function. Next, we need a \textit{decoder} to predict the output at a future time step, say $t+\tau$ given the temporal graph representation vector at time $t$; $\vy_{t+ \tau} = g(\vh_t)$. Architectures such as LSTM, GRU, TCN, ESN, and Transformer make a good choice for $f_{\rm T}(\cdot)$, whereas various GNN flavours can be used as $f_{\rm S}(\cdot)$ \cite{lai_lightcts_2023}. In the literature, TGNNs are often labelled as spatio-temporal GNNs \cite{jin_spatio-temporal_2023, wu_comprehensive_2021}. However, the term GNN inherently implies a focus on spatial learning.

In \cite{longa_graph_2023}, the authors have categorized TGNNs into two broad types, one based on model evolution, and the other based on embedding evolution. Here, they consider individual node representation $\vh_t^v$ for node $v \in \gV$ at time $t$. In \textit{model evolution} \cite{pareja_evolvegcn_2020}, the representation of $\gG_t$ can be evaluated by a function whose parameters evolve with time, i.e., $ \vh_t^v = f_{\rm S}\left(v, \gG_t; \bm{\Theta}_t \right)$, where the function parameter $\bm{\Theta}_t = f_{\rm T}\left( \bm{\Theta}_{t-1}, \cdots \bm{\Theta}_{t- \tau}  \right)$. In \textit{embedding evolution}, the representation at time $t$ is given by a function of the representations of previous graphs, i.e., $\vh_t^v = f_{\rm T}\Big( f_{\rm S}(v, \gG_t; \bm{\Theta}), \cdots f_{\rm S}(v, \gG_{t- \tau}; \bm{\Theta})  \Big)$. Majority of the TGNN models in the literature are based on embedding evolution \cite{tortorella_dynamic_2021, you_roland_2022, niknam_dyvgrnn_2023, liu_graph-based_2023}.

\section{Classical Time Series Forecasting}
\label{sec:classic}
Since we are dealing with temporal graphs, it is worth exploring classical forecasting techniques employed to deal with time series data \cite{montgomery_introduction_2015}. Moreover, temporal graphs can be viewed as correlated time series data, where the information of different nodes are correlated, with the edge weights representing the strength of correlation between them \cite{lai_lightcts_2023}.

\subsection{Linear Regression}
The response at time $t$, $y_t$ is a function of the variables $\{ x_t^1, \cdots x_t^k \}$, i.e., $y_t = f( x_t^1, \cdots x_t^k )$. The goal is to determine the mean response $\hat{y}_t = \mathbb{E}[ f(x_t^1, \cdots x_t^k )]$ based on a set of observations over a span of time $t \in \gT$. In linear regression, $y_t = \vw^\top \vx_t$, where $\vx_t^\top = \begin{bmatrix}
    1 & x_t^1 & \cdots & x_t^k
\end{bmatrix}$ and $\vw \in \mathbb{R}^{k+1}$ is the parameter. To get the mean response we obtain $\hat{\vw}$ such that the sum of the mean square distance between the actual response and the mean estimate is minimised for the given set of observations.
\begin{align}
    \hat{\vw} = (\mX^\top \mX)^{-1} \mX^\top \vy; \quad \mX^\top = \begin{bmatrix}
        \vx_1&
        \cdots &
        \vx_T
    \end{bmatrix}, 
    \vy^\top = \begin{bmatrix}
        y_1&
        \cdots &
        y_T
    \end{bmatrix}. 
\end{align}

\subsection{Exponential Smoothing}
Consider the model $y_t = w + \epsilon_t$, with $y_t, w, \epsilon_t \in \mathbb{R}$, where $w$ does not change with time, and $\epsilon_t$ is the error uncorrelated with $w$, and changes with time. Smoothing techniques help visualise the pattern in a time series data by removing the sharp changes, resulting in a smooth curve that can describe an approximate trend. The error at time $t \in [T]$, is discounted by a factor $\theta^{T-t}$. Let $\bm{\theta} = [\theta^T \, \theta^{T-1} \,  \cdots \, 1]^\top$, and $\vy = [y_1 \, y_2 \, \cdots \, y_T]^\top$, then the estimator $\hat{w}$ is given by: $\hat{w} = (1-\theta) \bm{\theta}^\top \vy $.
The act of discounting the previous errors this way is referred to as first-order exponential smoothing. For $n$-order exponential smoothing, the model is defined as $y_t = \sum_{k=0}^{n} w_k \frac{t^k}{k!} \, + \epsilon_t$. The goal is to find the set of parameters $\{ \hat{w}_k : k \in [n]\}$ to perform forecasting.

\subsection{Autoregressive Models}
In this section, we will explore the famous autoregressive integrated moving average (ARIMA) and vector autoregressive (VAR) models, which find extensive application in time series forecasting.
In autoregressive models, the variable at time $t$ is defined as the regression of the preceding variables. 

\begin{definition}[ARIMA]
    The ARIMA~$(p,d,q)$ model is given by:
    \begin{align}
        y_t = \delta + \epsilon_t + \sum_{k=1}^{p+d} \phi_k y_{t-k} - \sum_{k=1}^{q} \theta_k \epsilon_{t-k},
    \end{align}
    where $\delta$ is the uncorrelated component, $\epsilon_t$ is the noise at time $t$ and $\psi_k$ and $\theta_k$ are weights.
\end{definition}

\begin{definition}[VAR]
    The VAR~$(p)$ model is given by $\vy_t = \bm{\delta} + \sum_{k=1}^{p} \bm{\Phi}_k \vy_{t-k} + \bm{\epsilon}_t$.
\end{definition}

\begin{definition}[Forecast]
    The $\tau$ look-ahead forecast is defined as  $\hat{\vy}_{t+\tau} = \mathbb{E}\big[ \vy_{t+\tau} |\, \vy_{t}, \vy_{t-1}, \cdots \big]$ for $y_t \in \mathbb{R}^m$, $m \in \mathbb{N}$.
\end{definition}

\subsection{Bayesian Filters}
In Bayesian inference \cite{sarkka_bayesian_2023}, the goal is to model the system variables as random variables. As the name suggests, Bayesian inference relies heavily on Bayes' rule, where the posterior distribution of a parameter $\theta$, given the data $\gY = \{ y_1, y_2, \cdots y_t \}$ is written as: 
\begin{align}
    p(\theta | \gY) = \frac{p(\theta) p( \gY | \theta)}{p(\gY)},
\end{align}
 where $p(\theta)$ is the prior, $p(\gY | \theta)$ is the likelihood, and $p(\gY) = \int_{\theta \in \Theta} p(\theta) p(\gY | \theta) \, d\theta$ is the evidence.

To obtain the maximum likelihood estimate (ML) of the parameter, we maximize the likelihood function with respect to the parameter, i.e., $\hat{\theta}_{\rm ML} = \arg \max_{\theta \in \Theta} \prod_{y \in \gY} p(y | \theta)$.
To obtain the maximum a posteriori (MAP) estimate, we maximize the posterior distribution, i.e., $\hat{\theta}_{\rm MAP} = \arg \max_{\theta \in \Theta} p(\theta | \gY)$.

\begin{definition}[Probabilistic state space model]
    Consider a state sequence $\{ \vh_t : \forall t \in \gT \}$, and observation sequence $\{ \vx_t : \forall t \in \gT \}$, where $\vh_t \in \mathbb{R}^l$, and $\vx_t \in \mathbb{R}^d$. Then the probabilistic state space model is defined as $\vx_t \sim p( \vx_t | \vh_t)$ and $\vh_t \sim p( \vh_t | \vh_{t-1})$, where the current observation is dependent on the current state, and the current state depends on the previous state.
\end{definition}

A probabilistic state space model is also referred to as a non-linear filtering model. The central theme of Bayesian filtering is to derive the marginal distribution $p(\vh_t | \vx_t, \vx_{t-1}, \cdots)$, also called the filtering distribution of the state $\vh_t$ given the observations $\{ \vx_t, \vx_{t-1}, \cdots \}$. The distribution $p(\vh_t | \vx_{t-1}, \vx_{t-2}, \cdots)$ is called the predicted distribution which is derived using the Chapman-Kolmogorov equation: $p\big(\vh_t | \vx_{t-1}, \vx_{t-2}, \cdots \big) = \int p(\vh_t | \vh_{t-1} ) \, p(\vh_{t-1} | \vh_{t-2}, \cdots ) \, d\vh_{t-1}$. The filtering distribution having seen the new observation $\vx_t$ is updated through Bayes' rule. The Kalman filter is a linear Bayesian filtering model in which the predicted and filtering distributions are Gaussian \cite{sarkka_bayesian_2023}.

\section{Research Directions}
\label{sec:direction}
\subsection{Limitations}
The TGNNs in the literature suffer from the following limitations.

\subsubsection*{Expressiveness} Since TGNNs, rely on GNNs for spatial learning, they are limited by the expressiveness of the GNNs, i.e., the ability of the model to treat two isomorphic graphs as same. Graph isomorphism remains an unsolved problem as no algorithm exists which can say for sure that two graphs are isomorphic. Moreover, the graph isomorphism problem is NP unknown. However, it has been shown that GNNs are as expressive as the Weisfeiler-Lehman isomorphism test \cite{chen_weisfeiler-lehman_2023}.
\subsubsection*{Learnability} In deep learning, a deep model with large number of parameters is prone to overfitting in the absence of a well-crafted loss function. Moreover, a good learning strategy is essential to prevent the model parameters from being stuck in a local extremum.
\subsubsection*{Interpretability} The encoder-decoder architecture of the TGNNs involve learning a latent representation of the graph which is then processed to perform the downstream learning task. The latent vectors are intuitively
intractable and depend on the encoder, thereby reducing the interpretability of the TGL method. Taking inspiration from classical time series literature can help in crafting interpretable models.
\subsubsection*{Evaluation} In TGL, the models discern the inter-nodal correlations by exploiting the graph structure. However, not all graph datasets exhibit this property. Therefore, we need truly graphical datasets to be able to evaluate the efficacy of TGL techniques in a fair manner.
\subsubsection*{Scalability} In GNNs, each node aggregates the information from its nodes for improving the node representation. For a graph having $n$ nodes, a node can have up to $n-1$ neighbours. When $n$ becomes large, the learning becomes slow. To overcome this neighbour sampling is adopted, but that leads to poor expressiveness. Therefore methods need to be develop which are scalable and computationally light.
\subsubsection*{Privacy} The information about the input data can be inferred through model inversion attacks by analysing the model output or gradients. Moreover, in graph structured data, information pertaining to a node can be estimated from the neighbours. Therefore, even if we hide the sensitive information of some nodes, they are still at risk of being revealed. Research is being done on differential privacy methods on GNNs \cite{mueller_differentially_2023}, which can be extended to TGNNs.

\subsection{Applications}
Any data represented as a temporal graph, where the features of nodes or edges change over time, can be utilized for conducting regression or classification tasks on either nodes or edges.

\subsubsection*{Transportation} 
Traffic data is spatiotemporal data where the nodes represent sensors, and the edges are related to the geographical distance between any two sensors, while the node features can be flow count or flow speed. Such graphical data can be used to perform a variety of tasks such as \textit{traffic flow prediction}~\cite{fang_spatial-temporal_2021}, \textit{traffic speed prediction}~\cite{li_diffusion_2017}, and \textit{traffic congestion prediction}~\cite{di_traffic_2019}. The predicted traffic data can be further analyzed for \textit{route suggestion}~\cite{yin_deep_2022} to alleviate traffic congestion during peak hours. On-demand taxi services can use predicted traffic data for \textit{dynamic pricing}~\cite{he_spatio-temporal_2019}.

\subsubsection*{Weather} Weather data is collected through sensors located geographically, so the edges correspond to the geographical distance between sensor nodes, while the node features can represent weather data such as temperature, precipitation level, air quality, wind speed, and UV index. The weather data collected can be used to perform \textit{air quality inference}~\cite{cheng_neural_2018}, \textit{precipitation prediction}~\cite{li_using_2022}, \textit{wind speed prediction}~\cite{khodayar_spatio-temporal_2019}, and \textit{extreme weather prediction}~\cite{ni_ge-stdgn_2022}.

\subsubsection*{Infrastructure} The data from power grids can be used to \textit{predict energy consumption demand}~\cite{song_powercast_2017}. Similarly, data from water supply monitoring sensors can be used to \textit{predict water consumption demand}~\cite{donkor_urban_2014}. A similar approach can be adopted for natural resources like oil and gas to predict demand and supply.

\subsubsection*{Finance} In a stock market, a group of stocks can be forcibly presented as a graph with the node features being stock price and other details related to the stock. The edges between two stocks can be determined as a function of the stock features. As the stock prices evolve, the data becomes a temporal graph that can be analyzed to perform \textit{stock price prediction}~\cite{li_pearson_2022, chen_incorporating_2018}.

\subsubsection*{Social Network} In a social network, users are like nodes; the connections between them are the edges. The node features can capture details about user activity. This setup is studied in opinion dynamics~\cite{acemoglu_opinion_2011}, where the graph data is used for \textit{sentiment analysis}~\cite{pozzi_sentiment_2016}, where we analyze how users feel about a topic. Additionally, users are grouped into communities based on similarities in their activities; this is called \textit{community detection}~\cite{bedi_community_2016}. In the temporal setting, we can learn \textit{sentiment evolution}, \textit{community evolution}, and \textit{information diffusion}. The temporal activity of specific users can also be analyzed to \textit{identify bot accounts}~\cite{varol_online_2017} on a social media platform.

\subsubsection*{Supply Chain} In a supply chain, the producers, distributors, warehouses, and retailers can be depicted as nodes, while the edges can represent the connection between the different entities. The node features can represent the details related to a particular item, such as its price, quantity, and expiry date, to name a few. This temporal graph can then be analyzed to perform \textit{demand and supply forecast}~\cite{valles-perez_approaching_2022}, and therefore enable \textit{smart inventory management}, optimize \textit{logistic planning}, and perform \textit{dynamic pricing} of the items.

\subsubsection*{Communication Network} A wireless network can be naturally represented as a graph with the users as nodes and the channels as edges. The node feature can encapsulate spectrum allocation, data rate, transmit power, noise level, and data traffic demand. As the node features evolve, we get a temporal graph that can be analyzed to perform \textit{dynamic resource allocation}~\cite{eisen_optimal_2020} to optimize the data rate of all the users collectively.

\subsubsection*{Epidemiology} In epidemiology, the nodes can represent geographical locations such as cities, and their features can be the number of people infected, number of recovered patients, fraction of population vaccinated, and other details. Moreover, the edge features can be the number of travelers between any two cities through different modes of transport. This temporal graph can then be analyzed to \textit{predict the spread of the infection}~\cite{panagopoulos_transfer_2021} across different locations and aid in formulating preventive measures.

\subsubsection*{Neuroscience} The brain imaging data collected through techniques like EEG, MEG, fMRI, and fNIRS can be converted to temporal graph data, where the nodes represent different parts of the brain, and the neural activity level can be node features. The analysis of such temporal graphs can enable \textit{brain function classification}~\cite{bessadok_graph_2023} and \textit{diagnosis}~\cite{klepl_eeg-based_2022}.


\bibliographystyle{ACM-Reference-Format}
\bibliography{localrefs}




\end{document}